\documentclass[a4paper, onecolumn, 11pt]{article}
\usepackage[top=1in, bottom=1in, left=1in, right=1in]{geometry}
\usepackage{cite}
\usepackage{amsmath,amsthm,amssymb,amsfonts}
\usepackage{palatino}
\usepackage{bm}
\usepackage[colorlinks,
            linkcolor=cyan,
            anchorcolor=cyan,
            urlcolor=cyan,
            citecolor=cyan
]{hyperref}
\usepackage{graphicx}
\usepackage{caption}
\usepackage{subcaption} 
\usepackage{float}
\graphicspath{{./images/}}

\usepackage{array}
\usepackage{booktabs}
\usepackage{authblk}
\usepackage{multirow}
\usepackage{microtype}
\title{Seeing the Heat: Synthesizing High-Resolution Wood Thermal Responses from Optical Imagery}

\author{Jingren Xie}
\date{}

\begin{document}
\maketitle

\begin{abstract}
\noindent
The thermal behavior of wood is a critical factor in advanced material assembly.
However, pixel-level thermal analysis remains fundamentally constrained by the low resolution and noise inherent to infrared thermography.
To address this, we introduce an end-to-end computational framework that synthesizes high-resolution thermal responses directly from wood RGB images.
We first establish a core physical linkage: because spatial color variation in natural wood is driven by cellular anatomy, optical intensity serves as a reliable geometric proxy for the localized solid volume fraction.
By leveraging this theoretical insight, we develop an automated finite-element-method data engine that maps pixel-level optical intensity to a 3D thermodynamic voxel grid, generating high-fidelity synthetic thermal responses. 
We find that 1) when the thermal conductivity along the thickness direction is uniform or linear, wood RGB images and their corresponding thermal responses exhibit extreme morphological similarities, and the lateral thermal diffusion acts as a low-pass filter that smooths out high-frequency details; 2) when the thermal conductivity along the thickness direction is random, such morphological similarities are destroyed, and wood's 3D structure dominantly governs its thermal response.
We further utilize these synthetic thermal responses to supervise a neural surrogate model built upon the DINOv3 foundation model. 
Our results demonstrate that the neural surrogate model successfully internalizes the governing thermodynamic laws, thereby bypassing computationally expensive simulations and enabling high-resolution thermal inference. 
This methodology effectively bridges the semantic and thermodynamic domains, unlocking systematic, pixel-level analysis of fine-grained wood thermal responses.
Project: \color{cyan}{https://zekifayes.github.io/seeheat}
\end{abstract}

\section{Introduction}
The thermal behavior of wood plays a crucial role in advancing wood material assembly strategies~\cite{menges2015performative, tibbits2016self, cheng2020multifunctional, fragkia2020wood, fragkia2023thermodynamic}. 
While these strategies hold significant potential, they currently remain largely conceptual. 
Furthermore, pixel-level analysis of wood thermal behavior remains fundamentally unexplored. 
Although experimental setups with infrared cameras can capture low-resolution, noisy thermal data,
it is challenging to obtain high-fidelity, high-resolution thermal measurements. 
The absence of such high-quality data impedes a systematic, fine-grained analysis of wood thermal responses.
Meanwhile, foundational knowledge and insight into wood thermal behavior remain largely lacking.

This study is motivated by the empirical observation that wood RGB images and their corresponding thermal responses exhibit discernible morphological similarities and an inverse relationship between optical intensity and temperature difference~\cite{xie2026physics}. 
This morphological alignment suggests a fundamental physical linkage between a wood sample's optical intensity and its localized thermodynamic behavior. 
Because spatial color variation in uncoated, natural wood is primarily governed by cellular anatomy~\cite{marschner2005measuring}, optical intensity can serve as a geometric proxy for the localized solid volume fraction.

\begin{figure}[t]
    \centering
    \includegraphics[width=1.0\textwidth]{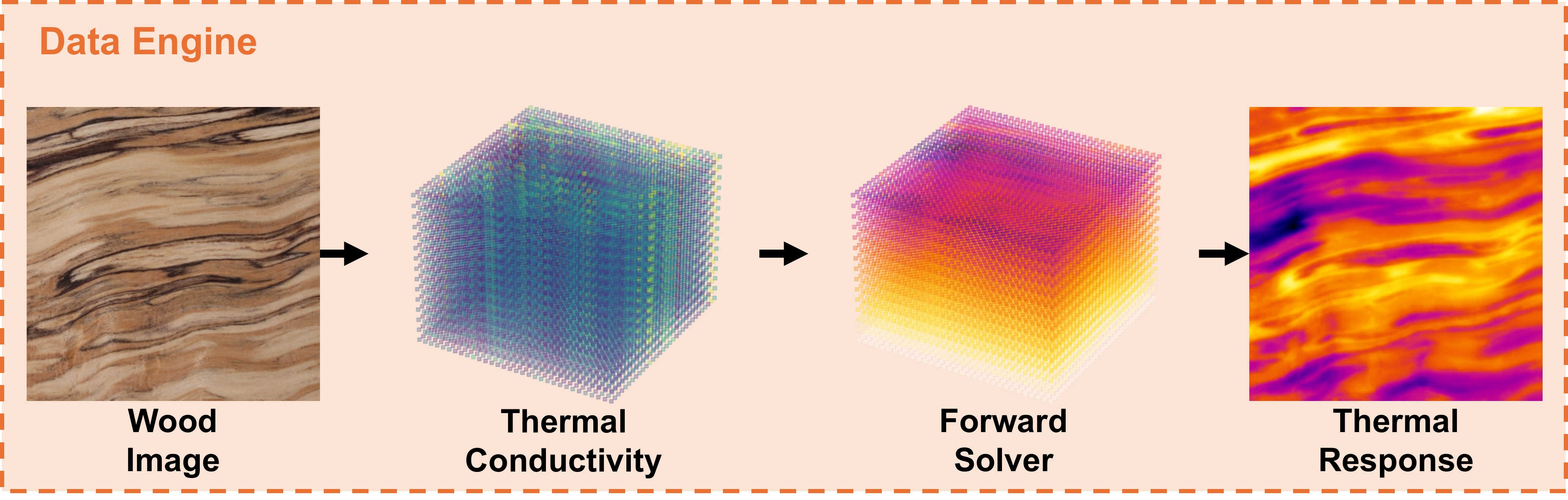}
    \caption{
    Overview of the Visual-Thermal Computational Framework. 
    The \textbf{Data Engine} acts as a physics-based solver, mapping a wood RGB image to a thermal conductivity tensor and solving Eq.~\ref{eq:3d_anisotropic_general} to produce a high-fidelity thermal response. 
    }
    \label{fig:overview_pipeline}
\end{figure}

However, it presents a significant methodological challenge to translate these complex, high-frequency optical patterns into continuous thermal responses. 
An intuitive solution is to use generative models (such as Generative Adversarial Networks~\cite{goodfellow2014generative}, Variational Autoencoders~\cite{kingma2013auto}, diffusion models~\cite {ho2020denoising, song2020score}, and flow models~\cite{lipman2022flow, liu2022flow}) to synthesize high-resolution thermal responses based on noisy, low-resolution thermal data.
However, these models are physically agnostic and fully supervised.
To bridge this gap, this paper proposes an end-to-end computational framework to synthesize high-resolution thermal responses from wood RGB images (Fig.~\ref{fig:overview_pipeline}).
We develop a physics-based data engine: we leverage optical intensity to estimate local structural density, map it to the physical thermal conductivity, and generate high-fidelity surface thermal responses via steady-state heat conduction simulations.
We further utilize this synthesized paired dataset to supervise several DINOv3-based neural decoders. 
These neural surrogate models learn to bypass the computationally expensive finite element simulations, enabling the pixel-level prediction of fine-grained thermal responses.

The primary contributions of this work are summarized as follows:
1) \textbf{Visual-Thermal Data Engine:} We introduce an automated, physics-based finite-element-method data engine that bridges the visual and thermodynamic domains. 
By utilizing optical intensity as a reliable geometric proxy for the local solid volume fraction, this data engine maps wood RGB imagery to a 3D thermodynamic voxel grid, successfully synthesizing high-fidelity thermal responses.
2) \textbf{Neural Surrogate Modeling:} We develop a fast neural surrogate framework utilizing the DINOv3 foundation model to predict pixel-level thermal responses from wood RGB images, effectively bypassing computationally expensive numerical simulations. 
Through systematic evaluation of decoders, we demonstrate that Patch-to-Pixel Refiners outperform standard upsampling by reintroducing high-frequency visual geometry, anchoring smooth semantic tokens to sharp cellular boundaries.

The remainder of the paper is organized as follows: Section~\ref{sec:related_work} introduces the related work.
Section~\ref{sec:wood_modeling} describes wood thermodynamic modeling.
Section~\ref{sec:framework} details our visual-thermal computational framework.
Section~\ref{sec:prediction} elaborates on our wood thermal prediction framework.
Section~\ref{sec:experiments} shows our systematic evaluation of the proposed frameworks, and Section~\ref{sec:conclusion} makes concluding remarks and discusses future work.
 
\section{Related Work}
\label{sec:related_work}  
\paragraph{Visual Representation Learning.}
Recent advances in self-supervised visual representation learning have yielded general-purpose, pre-trained models capable of generalizing across a broad range of downstream applications, including complex image generation~\cite{shi2025latent, shi2025svg, zheng2025diffusion}.
By leveraging massive, diverse datasets, architectures such as DINO series~\cite{caron2021emerging, oquab2024dinov, simeoni2025dinov3} and masked autoencoders~\cite{he2022masked} learn robust, patch-level visual representations without the need for human annotation. 
Similarly, multimodal models like CLIP~\cite{radford2021learning} and SigLIP~\cite{zhai2023sigmoid, tschannen2025siglip} align visual and textual representations through contrastive objectives, enabling powerful zero-shot generalization across vision-language domains. 
These foundation models excel at capturing deep geometric and structural context from optical imagery.
Because models like DINOv3 are optimized for visual semantics through self-supervised proxy tasks, their dense feature spaces are inherently detached from thermodynamic reality.
This paper addresses this gap by systematically investigating the capacity of DINOv3~\cite{simeoni2025dinov3} features to regress complex, high-frequency thermal responses, demonstrating their generalization potential in the highly constrained domain of computational thermodynamics.

\section{Wood Thermodynamic Modeling}
\label{sec:wood_modeling}
Without internal heat generation, general 3D steady-state heat conduction within a wood sample is mathematically described using a partial differential equation (PDE)~\cite{Cengel2006}
\begin{equation}
\frac{\partial}{\partial x} \left( k_{x} \frac{\partial T}{\partial x} \right) +
\frac{\partial}{\partial y} \left( k_{y} \frac{\partial T}{\partial y} \right) +
\frac{\partial}{\partial z} \left( k_{z} \frac{\partial T}{\partial z} \right) = 0,
\label{eq:3d_anisotropic_general}
\end{equation}
where $T=T(x,y,z)$ is the temperature field within a spatial domain of $L_x \times L_y \times L_z$ (where $L_x=L_y \gg L_z$), and $k_x=k_x(x,y,z)$, $k_y=k_y(x,y,z)$, and $k_z=k_z(x,y,z)$ are the spatially-varying thermal conductivities along the $x$, $y$, and $z$ directions, respectively. 
Because the wood sample directly contacts the testbed, it satisfies a Dirichlet boundary condition
\begin{equation}
    T(x, y, z=0)=T_{\text{bed}}(x,y).
\end{equation}
Heat exchange via heat convection between the top surface and the ambient air satisfies a Robin boundary condition
\begin{equation}
    \left . -k_{\text{top}}  \frac{\partial T} {\partial z} \right|_{z=L_z} = h \left( T_{\text{top}} - T_{\infty} \right),
\end{equation}
where $k_{\text{top}} = k_z(x, y, L_{z})$ is the thermal conductivity on the top surface, $h$ is the convective coefficient, $T_{\text{top}} = T(x, y, L_{z})$ is the temperature distribution on the top surface, and $T_{\infty}$ is the ambient temperature. 
The remaining lateral surfaces satisfy Neumann boundary conditions
\begin{equation}
    \left. \frac{\partial T} {\partial x} \right |_{x=0, L_x}=0,
\left. \frac{ \partial T} {\partial y} \right |_{y=0, L_y}=0.
\end{equation}
More information is provided in Appendix~\ref{sec:verification}.

\section{Visual-Thermal Computational Framework}
\label{sec:framework}

This section provides a computational pipeline for synthesizing high-resolution wood thermal responses.
More information is provided in Appendix~\ref{sec:modeling}.

\subsection{Visual-Thermal Conductivity Mapping}
To derive an analytical solution for the thermal response, the thermal conductivities ($k_x$, $k_y$, and $k_z$) are assumed to be locally uniform within a small micro-neighborhood to factor out of the spatial derivatives.
Since we only have 2D image information rather than 3D structure information, for a single pixel $(x,y)$, $k_z$ remains uniform along the $z$-direction but varies across the $xy$-plane. 
By defining the anisotropy ratios as global constants ($\alpha = k_x/k_z$ and $\beta = k_y/k_z$) and setting $\alpha = \beta$ for simplicity, Eq.~\ref{eq:3d_anisotropic_general} reduces to
\begin{equation}
    \alpha \nabla^2 T + \frac{\partial^2 T}{\partial z^2} = 0.
\label{eq:3d_simpler}
\end{equation}
To account for lateral thermal diffusion $\alpha \nabla^2 T$, we apply a standard 2D Gaussian kernel $G_{\sigma}$ to approximate the surface thermal response
\begin{equation}
    \Delta T(x,y) \approx G_{\sigma} \ast \left( \frac{L_z q}{k_{\text{top}}(x,y)} \right),
    \label{eq:blurred_resistance}
\end{equation}
where $\Delta T(x,y) = T_{\text{bed}} - T_{\text{top}}$ is the temperature difference and $q = h(T_{\text{top}} - T_{\infty})$ is the surface convective heat flux.

In natural wood samples, spatial color variation is mainly governed by cellular anatomy~\cite{marschner2005measuring}.
Wood is a biological cellular solid comprising solid cell walls and air voids. 
The interfaces between the solid cell walls and air voids create significant refractive index mismatches, leading to intense light scattering~\cite{vasileva2018light}. 
Highly porous regions scatter more ambient light, resulting in bright pixels~\cite{kitamura2016determination}. 
Dense regions have fewer scattering interfaces, leading to greater photon absorption and darker pixels~\cite{vasileva2018light}. 
Therefore, after normalizing the wood RGB image into a monochromatic map $I(x,y) \in [0, 1]$, the term $(1-I)$ serves as a photometric density proxy for the solid volume fraction $V_f$ of the local cellular matrix, $V_f(x,y) \approx 1 - I(x,y)$.
The effective thermal conductivity of a two-phase cellular solid is governed by the Gibson-Ashby model~\cite{GibsonAshby1997}. 
Because the thermal conductivity of the solid cell wall material ($0.34$ W/(m·K)) is far greater than that of the internal air ($0.026$ W/(m·K))~\cite{zhao2023scalable}, $k_{\text{top}}$ is approximately proportional to $V_f$. 
This yields $k_{\text{top}}(x,y) \approx k_{\max} \cdot V_f(x,y)$, where $k_{\max}$ represents the theoretical intrinsic thermal conductivity of a completely solid, void-free pixel.
Substituting the photometric density proxy establishes a linear mathematical relationship between optical intensity $I$ and thermal conductivity $k_{\text{top}}$
\begin{equation}
k_{\text{top}}(x,y) = k_{\max} \left[ 1 - I(x,y) \right].
\label{eq:photometric_proxy}
\end{equation}
We invert Eq.~\ref{eq:blurred_resistance} and apply a first-order Taylor approximation to yield
\begin{equation}
    \frac{1}{\Delta T} \approx \eta \left[ G_{\sigma} \ast (1-I) \right],
    \label{eq:w_parameter}
\end{equation}
where $\eta = k_{\max} / (L_z q)$ is a system parameter.
We approximate the macroscopic heat flux $q$ using 1D Fourier's law over the bulk wood sample, $q \approx \bar{q} = \bar{k}_{\text{bulk}} \frac{\Delta \bar{T}}{L_z}$, where $\bar{k}_{\text{bulk}}$ is the macroscopic bulk thermal conductivity and $\Delta \bar{T}$ is the spatially averaged temperature difference. 
Comparing the parameters yields $k_{\max} = \eta \bar{k}_{\text{bulk}} \Delta \bar{T}$. 
For simplicity, we non-dimensionalize the macroscopic thermal potential by normalizing $\bar{k}_{\text{bulk}} \Delta \bar{T} = 1$. 
The system parameter $\eta$ represents the effective intrinsic thermal conductivity ($\eta = k_{\max}$)
\begin{equation}
k_{\text{top}}(x,y) = \eta \left[ 1 - I(x,y) \right].
\label{eq:final_synthesis}
\end{equation}

\begin{figure}[t]
    \centering
    \includegraphics[width=1.0\textwidth]{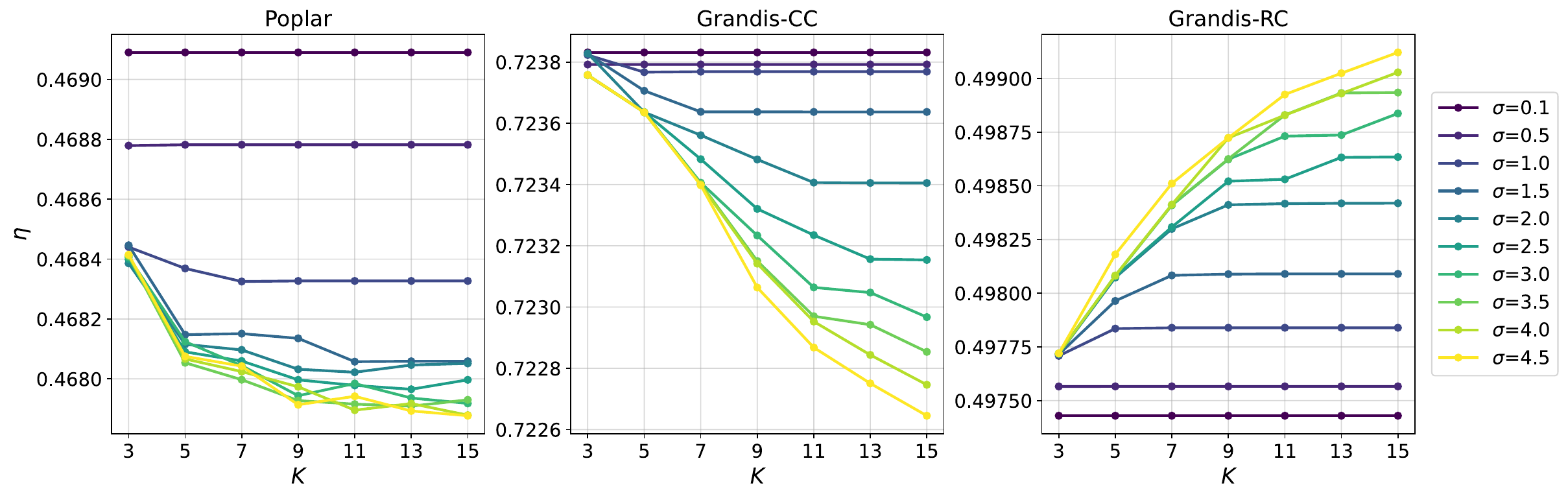}
    \caption{
    Stability of the system parameter $\eta$ across multiscale spatial configurations.
    The parameter $\eta$ is optimized using MAE across various $K$ and $\sigma$ for the Poplar, Grandis-CC, and Grandis-RC datasets. 
    The absolute value of $\eta$ remains tightly constrained within a narrow numerical band for each dataset, demonstrating high model stability.
    }
    \label{fig:w_param}
\end{figure}

\paragraph{Numerical Stability of the System Parameter $\eta$.}
Based on the real-world thermal datasets in ~\cite{xie2026physics}, we regress $\eta$ using experimental thermal responses optimized via a Mean Absolute Error (MAE) objective function.
The Gaussian kernel parameters span sizes $K=\{3, 5, 7, 9, 11, 13, 15\}$ and variances $\sigma = \{ 0.1, 0.5, 1.0, \dots, 4.5 \}$.
As illustrated in Fig.~\ref{fig:w_param}, $\eta$ exhibits remarkable stability. 
While it slightly decreases for Poplar ($\eta \approx 0.468$) and Grandis-CC ($\eta \approx 0.723$) and increases for Grandis-RC ($\eta \approx 0.498$) with kernel growth, the total variation is confined to a tight numerical band on the order of $10^{-3}$. 
This strict constraint serves as an intrinsic, robust visual-thermal property invariant under the underlying multiscale image-processing parameters.

\subsection{Visual-Thermal Computational Framework}
\label{sec:synthetic_pipeline}

To generate high-resolution thermal responses, we develop an automated pipeline based on the Finite Element Method (FEM). 
This visual-thermal computational framework bridges the discrete geometric domain of the optical intensity with a 3D thermodynamic solver (Fig.~\ref{fig:overview_pipeline}).
More information is provided in Appendix~\ref{sec:derivation_weak_form}.

\paragraph{Domain Discretization and Material Mapping.}
The wood sample is reconstructed as a 3D structured cuboid mesh with dimensions $L_x \times L_y \times L_z$. 
To prevent spatial interpolation errors, the mesh resolution $n_x \times n_y$ is defined to perfectly match the high-resolution pixel dimensions of the input wood RGB image.
Because the wood sample operates within a thin-plate regime ($L_z \ll L_x, L_y$), the resolution $n_z$ utilizes stretched elements to optimize computational efficiency.
The 2D optical proxy $1-I(x,y)$ is computationally projected down the $z$-axis, formulating the fully anisotropic thermal conductivity tensor $\mathbf{K}$
\begin{equation}
    \mathbf{K}(x,y) = 
    \begin{bmatrix} 
    \alpha \cdot k_{\text{top}}(x,y) & 0 & 0 \\ 
    0 & \beta \cdot k_{\text{top}}(x,y) & 0 \\ 
    0 & 0 & k_{\text{top}}(x,y) 
    \end{bmatrix}.
\end{equation}

\paragraph{Boundary Value Problem Formulation.}
Eq.~\ref{eq:3d_anisotropic_general} is assembled into its weak variational form. 
Let $V$ represent the continuous Galerkin function space. 
We seek a temperature field $T \in V$ such that for all test functions $v \in V$, the system satisfies
\begin{equation}
\int_{\Omega} (\mathbf{K} \nabla T) \cdot \nabla v \, d\Omega + \int_{\Gamma_{\text{top}}} h T v \, ds = \int_{\Gamma_{\text{top}}} h T_{\infty} v \, ds.
\end{equation}
The Dirichlet boundary condition is enforced as a strict nodal constraint at $z=0$, while the Robin boundary condition governing convective heat flux emerges as the integral over the top surface $\Gamma_{\text{top}}$. 
Given the vast degrees of freedom afforded by pixel-level spatial resolution, the resulting linear system is solved using a conjugate gradient method with an Incomplete LU preconditioner via FEniCS~\cite{alnaes2015fenics}. 
The data engine extracts the 2D surface slice $T_{\text{top}}$, serving as our final thermal response.

\begin{figure}[t]
    \centering
    \includegraphics[width=1.0\textwidth]{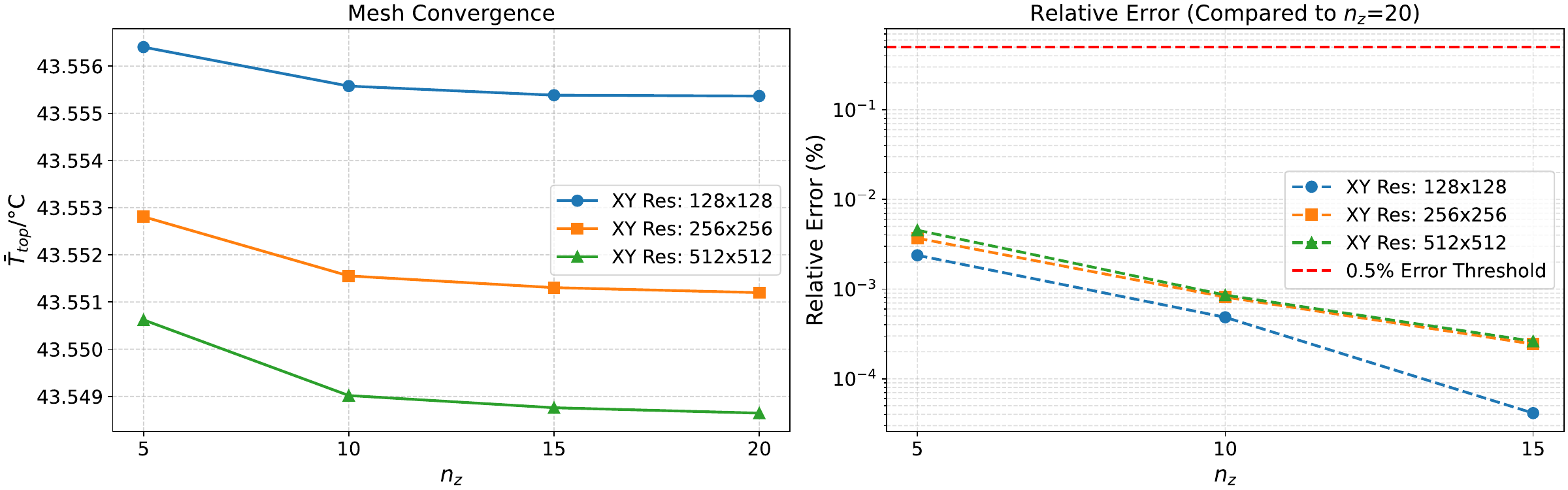}
    \caption{
    Mesh convergence analysis for the 3D finite element thermal simulation. 
    (Left) The calculated average top surface temperature $\bar{T}_{\text{top}}$ quickly stabilizes across different discretizations $n_z$ for varying high-resolution grids. 
    (Right) The relative error of each configuration, evaluated against the $n_z=20$ baseline, remains orders of magnitude below a conservative 0.5\% error threshold, validating the numerical stability of the thin-plate regime.
    }
    \label{fig:convergence}
\end{figure}

\paragraph{Mesh Convergence and Numerical Stability.} 
To validate the domain discretization strategy, a comprehensive mesh convergence study is performed (Fig.~\ref{fig:convergence}). 
The average simulated top surface temperature rapidly stabilizes as the number of layers $n_z$ increases.
The relative error along the $z$-axis, when evaluated against the finest baseline $n_z=20$, falls beneath $10^{-2}\%$. 
Because this deviation is several orders of magnitude below the strict $0.5\%$ tolerance threshold, the geometric model operates reliably within a stable thin-plate regime. 
The pipeline safely utilizes stretched elements (e.g., $n_z = 10$) to maximize efficiency.


\section{Wood Thermal Prediction}
\label{sec:prediction}

\begin{figure}[t]
    \centering
    \includegraphics[width=1.0\textwidth]{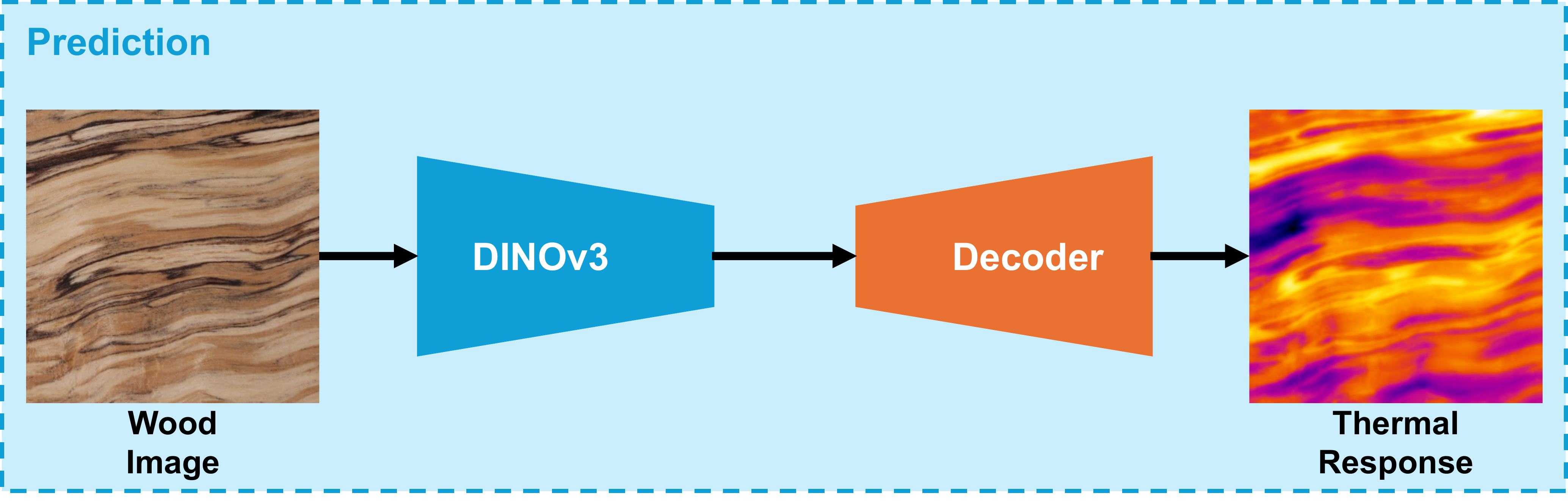}
    \caption{
    Overview of the Wood Thermal Prediction Framework. 
    It leverages this synthesized paired dataset to supervise a DINOv3-based encoder-decoder architecture, enabling direct prediction of the thermal responses from the wood RGB images.
    }
    \label{fig:mathematical_framework}
\end{figure}

While the FEM data engine generates accurate thermal responses, it is computationally expensive to compute these dense PDEs. 
To bypass this bottleneck, we leverage multi-scale semantic features extracted from pre-trained visual foundation models of varying scales (DINOv3~\cite{simeoni2025dinov3} small, base, and large: ViT-S, ViT-B, and ViT-L) to regress high-fidelity thermal responses from wood RGB images (Fig.~\ref{fig:mathematical_framework}). 
To explore how these rich, highly compressed semantic features can best be mapped to dense, high-resolution thermal responses, we design a progression of simple yet highly effective neural decoders. 

\subsection{Foundation Feature Extraction}
Given an RGB image $X \in \mathbb{R}^{3 \times H \times W}$, the frozen DINOv3 encoder partitions the RGB image into nonoverlapping patches of size $P \times P$. 
This results in a sequence of $N = (H / P) \times (W / P)$ patch tokens. 
To avoid heuristically selecting a single transformer layer, we fuse the multi-scale representations across all $L$ transformer layers using a learnable softmax function. 
Let $f_l \in \mathbb{R}^{N \times D}$ represent the token sequence extracted from the $l$-th transformer layer. 
The aggregated token sequence $z_{\text{seq}} \in \mathbb{R}^{N \times D}$ is computed as
\begin{equation}
z_{\text{seq}} = \sum_{l=1}^{L} \left( \frac{\exp(\theta_l)}{\sum_{j=1}^{L} \exp(\theta_j)} \right) f_l,
\end{equation}
where $\theta \in \mathbb{R}^L$ denotes the vector of learnable layer parameters, and $D$ is the embedding dimension. 
By unflattening the aggregated sequence back into its 2D spatial arrangement, we obtain the base semantic feature map $z_{\text{fuse}} \in \mathbb{R}^{D \times h \times w}$, where $h = H/P$ and $w = W/P$. 
$z_{\text{fuse}}$ serves as the foundational input for all subsequent decoding explorations.

\subsection{Decoders}
The objective is to design a fast and effective decoder.
We explore linear heads, linear heads with optical incorporation, and finally, direct patch-to-pixel refinement.

\paragraph{Linear Head (LH).}
We construct a naive baseline that directly maps the aggregated semantic features to the thermal response without any additional contextual refinement. 
The compressed semantic features are projected onto a single-channel fused map using a $1\times1$ convolution and are subsequently bilinearly interpolated back to the native image resolution
\begin{equation}
\hat{T}_{\text{top}} = \text{Interp}_{h \times w \to H \times W} \left( \text{Conv}_{1 \times 1}(z_{\text{fuse}}) \right).
\end{equation}

\paragraph{Two-Layer Head (TLH).}
To evaluate whether the token projection merely requires deeper pointwise capacity, we add a nonlinear activation (GELU) between two $1 \times 1$ convolutions to enhance the expressivity of the regression head before interpolation
\begin{equation}
\hat{T}_{\text{top}} = \text{Interp}_{h \times w \to H \times W} \left( \text{Conv}_{1\times1} \left( \text{GELU}\left(\text{Conv}_{1\times1}(z_{\text{fuse}})\right) \right) \right).
\end{equation}

\paragraph{Optical Incorporation Head (OIH).}
We utilize the identical semantic bottleneck to generate the interpolated base thermal map, but explicitly reintroduce the high-frequency boundaries of the original image. 
We concatenate the semantic map directly with the raw, 3-channel optical input $X$.
A $1\times1$ convolution fuses the combined tensor into the final thermal response
\begin{equation}
\hat{T}_{\text{top}} = \text{Conv}_{1 \times 1} \left( \left[ \text{Interp}_{h \times w \to H \times W} \left( \text{Conv}_{1 \times 1} (z_{\text{fuse}}) \right) \parallel X \right] \right).
\end{equation}

\paragraph{Patch-to-Pixel Refiner.}
Rather than upsampling a compressed single-channel map, we replace the spatial interpolation step by introducing a patch refiner ($\mathcal{R}$) that inverts the ViT tokenization process. 
Operating directly on the spatially arranged foundational grid $z_{\text{fuse}}$, the refiner projects the deep semantic vectors directly back into their corresponding dense pixel blocks. 
It expands the channel dimension to generate a subpixel feature map $\hat{z}_{\text{pixel}} \in \mathbb{R}^{(C_{\text{out}} \cdot P^2) \times h \times w}$, where $C_{\text{out}} = 1$ represents the thermal response
\begin{equation}
\hat{z}_{\text{pixel}} = \mathcal{R}(z_{\text{fuse}}).
\end{equation}
$z_{\text{fuse}}$ is then unpatched back into the high-resolution spatial domain using the standard subpixel convolution operator, PixelShuffle \cite{shi2016real}
\begin{equation}
\hat{T}_{\text{top}} = \text{PixelShuffle} \left( \hat{z}_{\text{pixel}} \right).
\end{equation}
By mapping the embedding dimension $D$ to $P^2$ independent pixels, we instantiate the refiner using two distinct architectural paradigms: Local and Global Contextual Refinement.

\paragraph{Local Contextual Refinement (R-CNN).}
To test whether tokens only require isolated, channel-wise domain adaptation to reconstruct their internal pixels, we implement the refiner using two pointwise $1\times 1$ convolutions
\begin{equation}
\mathcal{R}_{\text{CNN}}(z_{\text{fuse}}) = \text{Conv}_{1\times1} \left( \text{GELU}\left(\text{Conv}_{1\times1}(z_{\text{fuse}})\right) \right).
\end{equation}

\paragraph{Global Contextual Refinement (R-Attn).}
To test whether modeling macroscopic thermodynamic behavior across the entire wood sample before pixel unfolding improves reconstruction, we implement a global refiner using a multi-head self-attention block~\cite{vaswani2017attention}
\begin{equation}
z_{\text{attn}} = z_{\text{seq}} + \text{Softmax}\left( \frac{(z_{\text{seq}} W_Q)(z_{\text{seq}} W_K)^T}{\sqrt{d_k}} \right) \left( z_{\text{seq}} W_V \right),
\end{equation}
where $W_Q$, $W_K$, and $W_V$ denote the learnable projection matrices for the queries, keys, and values, respectively, and $d_k$ represents the dimension of the attention heads.
$z_{\text{attn}}$ is reshaped and projected to the expanded subpixel dimension via a $1\times 1$ convolution
\begin{equation}
\mathcal{R}_{\text{Attn}}(z_{\text{attn}}) = \text{Conv}_{1\times1} \left( \text{Reshape}_{N \to h \times w}(z_{\text{attn}}) \right).
\end{equation}

\section{Experiments and Results}
\label{sec:experiments}

\begin{table}[t]
\centering
\caption{Quantitative results of different decoders based on DINOv3 backbones (ViT-S, ViT-B, ViT-L). 
Metrics include MAE, RMSE, and $\delta_{01}$. $\delta_{01}$ is presented in percentage.}
\label{tab:master_performance_summary}
\small
\begin{tabular}{l ccc ccc ccc}
\toprule
& \multicolumn{3}{c}{ViT-S} & \multicolumn{3}{c}{ViT-B} & \multicolumn{3}{c}{ViT-L} \\
\cmidrule(lr){2-4} \cmidrule(lr){5-7} \cmidrule(lr){8-10}
Decoders & MAE ($\downarrow$) & RMSE ($\downarrow$) & $\delta_{01}$ ($\uparrow$) & MAE ($\downarrow$) & RMSE ($\downarrow$) & $\delta_{01}$ ($\uparrow$) & MAE ($\downarrow$) & RMSE ($\downarrow$) & $\delta_{01}$ ($\uparrow$) \\
\midrule
    LH & 0.1150 & 0.1562 & 97.87 & 0.1068 & 0.1459 & 98.40 & 0.1022 & 0.1434 & 98.33 \\
    TLH & 0.0840 & 0.1221 & 98.98 & 0.0839 & 0.1220 & 98.98 & 0.0848 & 0.1227 & 98.96 \\
    OIH & 0.0933 & 0.1266 & 99.02 & 0.0842 & 0.1150 & 99.40 & 0.0768 & 0.1068 & 99.47 \\
    \addlinespace
    R-CNN & 0.0702 & 0.0995 & 99.61 & 0.0558 & 0.0809 & 99.78 & 0.0556 & 0.0784 & 99.84 \\
    R-Attn & 0.0652 & 0.0915 & 99.73 & 0.0603 & 0.0851 & 99.78 & 0.0592 & 0.0829 & 99.79 \\
\bottomrule
\end{tabular}
\end{table}

\begin{figure}[t]
    \centering
    \includegraphics[width=1.0\textwidth]{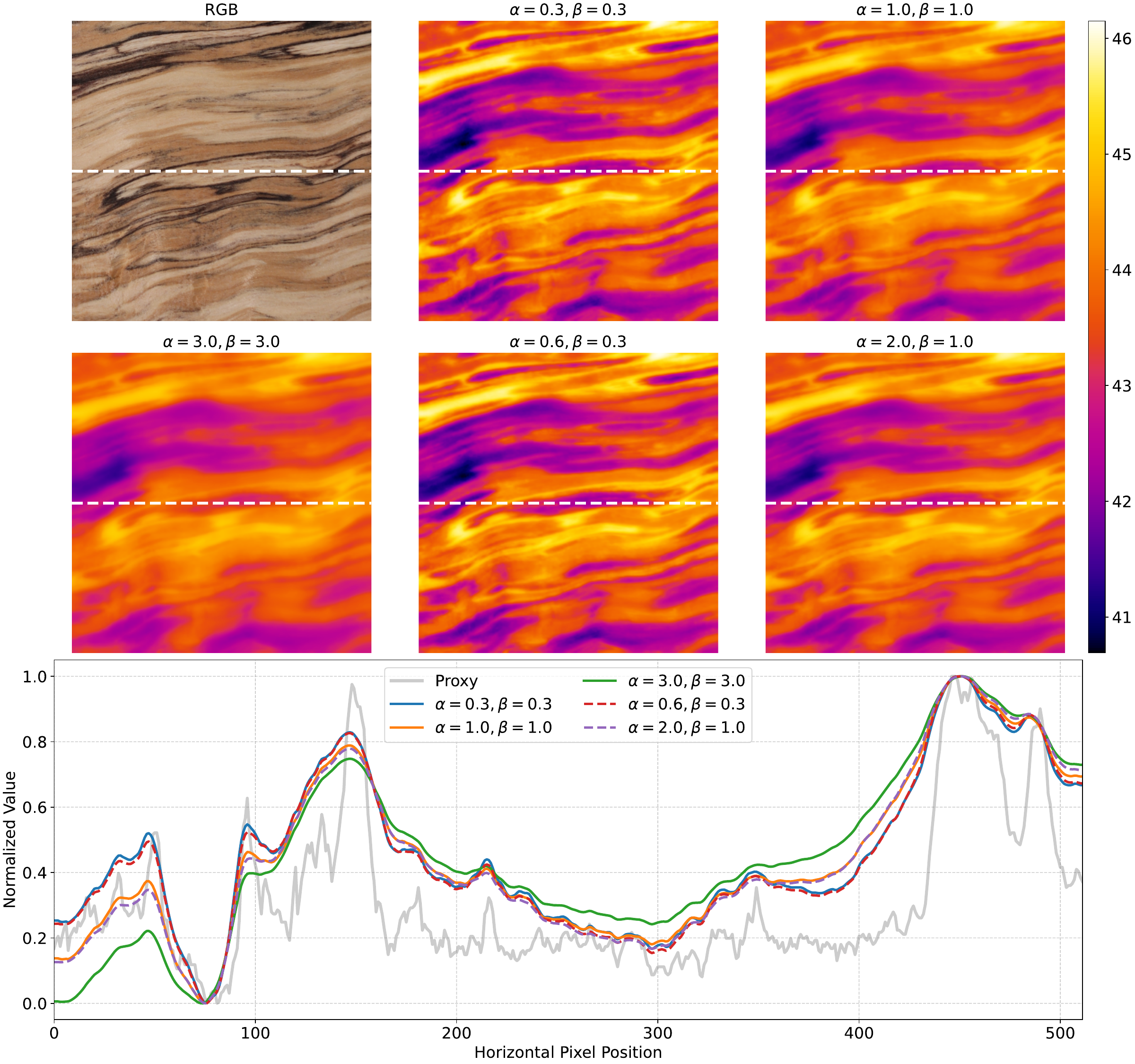}
     \caption{
     Parametric ablation study of thermal diffusion. 
     The top and middle rows display the original RGB image and the corresponding simulated steady-state surface temperature maps under various diffusion coefficients ($\alpha, \beta$).
     }
    \label{fig:lateral_diffusion}
\end{figure}

\begin{figure}[t]
    \centering
    \includegraphics[width=1.0\textwidth]{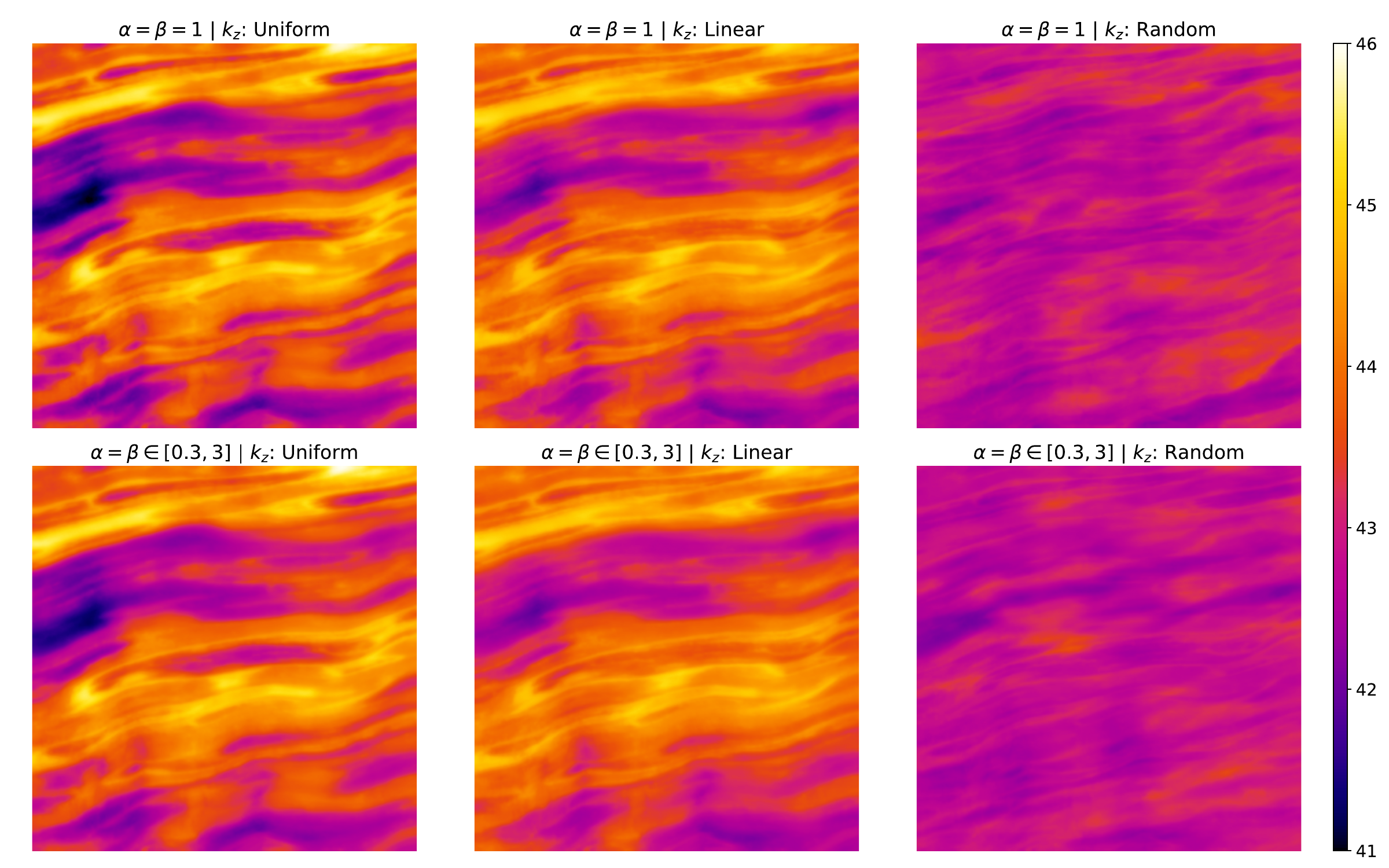}
     \caption{Wood thermal responses across six distinct structural configurations. 
     The top row demonstrates an idealized isotropic diffusion with coefficients $\alpha=\beta=1$, while the bottom row incorporates spatially varying in-plane anisotropy ($\alpha=\beta \in [0.3, 3]$). 
     The random $k_z$ distribution exhibits the lowest surface temperatures due to the thermal resistance introduced by heterogeneous layer stratification.}
    \label{fig:heat_2D}
\end{figure}

\begin{figure}[t]
    \centering
    \includegraphics[width=1.0\textwidth]{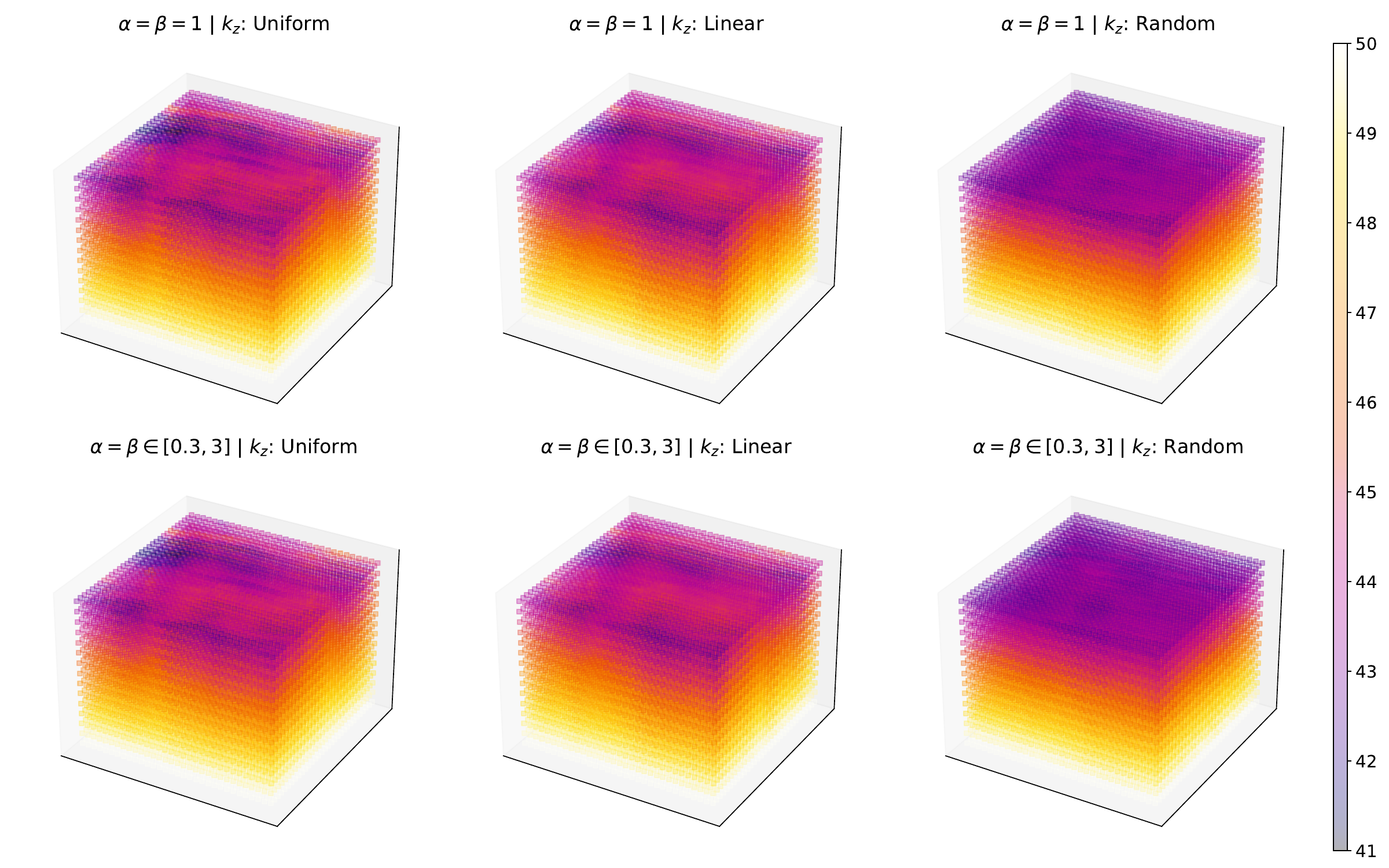}
     \caption{Wood internal heat transfer profiles mapping the temperature gradients from the bottom thermal boundary ($50^\circ\text{C}$) to the top surface.
     The uniform and linear $k_z$ distributions (left and center columns) facilitate relatively continuous vertical heat conduction. 
     In contrast, the randomly stratified distributions (right column) disrupt direct pathways, introducing tortuous thermal flow and non-linear temperature drops across the wood thickness.}
    \label{fig:heat_3D}
\end{figure}

\subsection{Experimental Setup}
\paragraph{Training Data.}
We evaluate our decoders on the synthetic dataset with a resolution of $512 \times 512$.
The dataset is partitioned into training, validation, and test sub-datasets in an 8:1:1 ratio.
There are a total of $1503$ paired RGB images and their corresponding thermal responses.

\paragraph{Training Details.}
All these models are trained on H100 using the Adam optimizer~\cite{kingma2014adam} with a batch size of $64$.
The learning rate is initially set to $0.01$ using an early-stopping mechanism.
The maximum iterations are set to $6000$.
Several data augmentations, including horizontal flipping, vertical flipping, and rotation, are applied to reduce overfitting during model training.

\paragraph{Evaluation Metrics.}
We employ three widely used metrics to evaluate the quality of thermal response prediction: 1) Mean Absolute Error,  $\text{MAE} = 1/HW \sum_{i=1}^{HW} |T_{i} - \hat{T}_{i}|$; 2) Root Mean Squared Error,  $\text{RMSE} = \sqrt{1/HW \sum_{i=1}^{HW} \left(T_{i} - \hat{T}_{i} \right)^2}$; 3) the percentage of pixels,  $\delta_{01} = \text{max}(T_{i}/\hat{T}_{i}, \hat{T}_{i}/T_{i}) < 1.01$.

\subsection{Results on Visual-Thermal Computational Framework}

\paragraph{Influence of lateral Thermal Diffusion.} 
We study the influence of the coefficients $\alpha$ and $\beta$ on the simulated thermal responses (Fig.~\ref{fig:lateral_diffusion}). 
Visual inspection reveals that lateral thermal diffusion intrinsically acts as a physical low-pass filter. 
Under restricted diffusion ($\alpha=0.3, \beta=0.3$), the thermal profile maintains sharp, high-frequency fidelity. 
Under large diffusion ($\alpha=3.0, \beta=3.0$), distinct localized thermal peaks are severely dampened, erasing the fine, structure-driven thermal gradients.
\textit{Finding: For any pixel $(x,y)$, the thermal conductivity $k_{z}$ is uniform along the $z$ direction but varies across the $xy$ plane; diffusion coefficients $\alpha$ and $\beta$ work as a low-pass filter to smooth out sharp, high-frequency details.}

\paragraph{Influence of Thermal Conductivity Distribution along the Thickness Direction.}
To evaluate the influence of the internal thermal conductivity distribution (Appendix~\ref{sec:thermal_conductivity_distributions}), we analyze the 2D thermal responses (Fig.~\ref{fig:heat_2D}) and the corresponding 3D internal heat transfer profiles (Fig.~\ref{fig:heat_3D}) across the simulated configurations. 
As observed in the 2D thermal responses, while the fundamental morphological patterns are dictated by the structural anatomy of the topmost layer, the underlying $k_z$ distribution profoundly modulates the overall temperature magnitude and gradient sharpness. 
The uniform and linear thermal conductivity distributions facilitate relatively continuous and predictable heat conduction, resulting in higher localized surface temperatures. 
The random thermal conductivity distribution---which represents a highly stratified, heterogeneous stacking of varying solid volume fractions---introduces tortuous thermal pathways. 
This internal stratification disrupts direct heat conduction, yielding the lowest overall surface temperatures and introducing complex, non-linear thermal drops from the bottom boundary ($50^\circ\text{C}$) to the top surface. 
Furthermore, the introduction of in-plane spatial anisotropy ($\alpha=\beta \in [0.3, 3]$) significantly alters localized lateral diffusion compared to the idealized isotropic baseline ($\alpha=\beta=1$), as shown in Figs~\ref{fig:heat_2D} and ~\ref{fig:heat_3D}.
\textit{Finding: when the thermal conductivity distribution along the thickness direction is uniform or linear, the wood RGB image and its thermal response exhibit great morphological similarities; the wood's 3D structure governs the thermal response when the thermal conductivity distribution along the thickness direction is random.}

\subsection{Wood Thermal Prediction}
\subsubsection{Main Results}
\paragraph{Numerical Configurations.}
The dimensions of the wood sample are $L_x=L_y=$ 100 mm and $L_z=$ 3 mm.
The spatial domain is discretized into $n_x \times n_y \times n_z = 512 \times 512 \times 10$.
The testbed temperature $T_{\text{bed}}$ is constrained to $50.0 ^\circ \mathrm{C}$, and the ambient temperature $T_{\infty}$ is set to $20.0 ^\circ \mathrm{C}$. 
The convective coefficient $h$ is configured to $20.0$ W/(m²·K), and $k_{\max}$ is bounded at $0.468$ W/(m·K). 
The lateral thermal diffusion coefficients are isotropic ($\alpha = \beta = 1$).

\begin{figure}[t]
    \centering
    \includegraphics[width=1.0\textwidth]{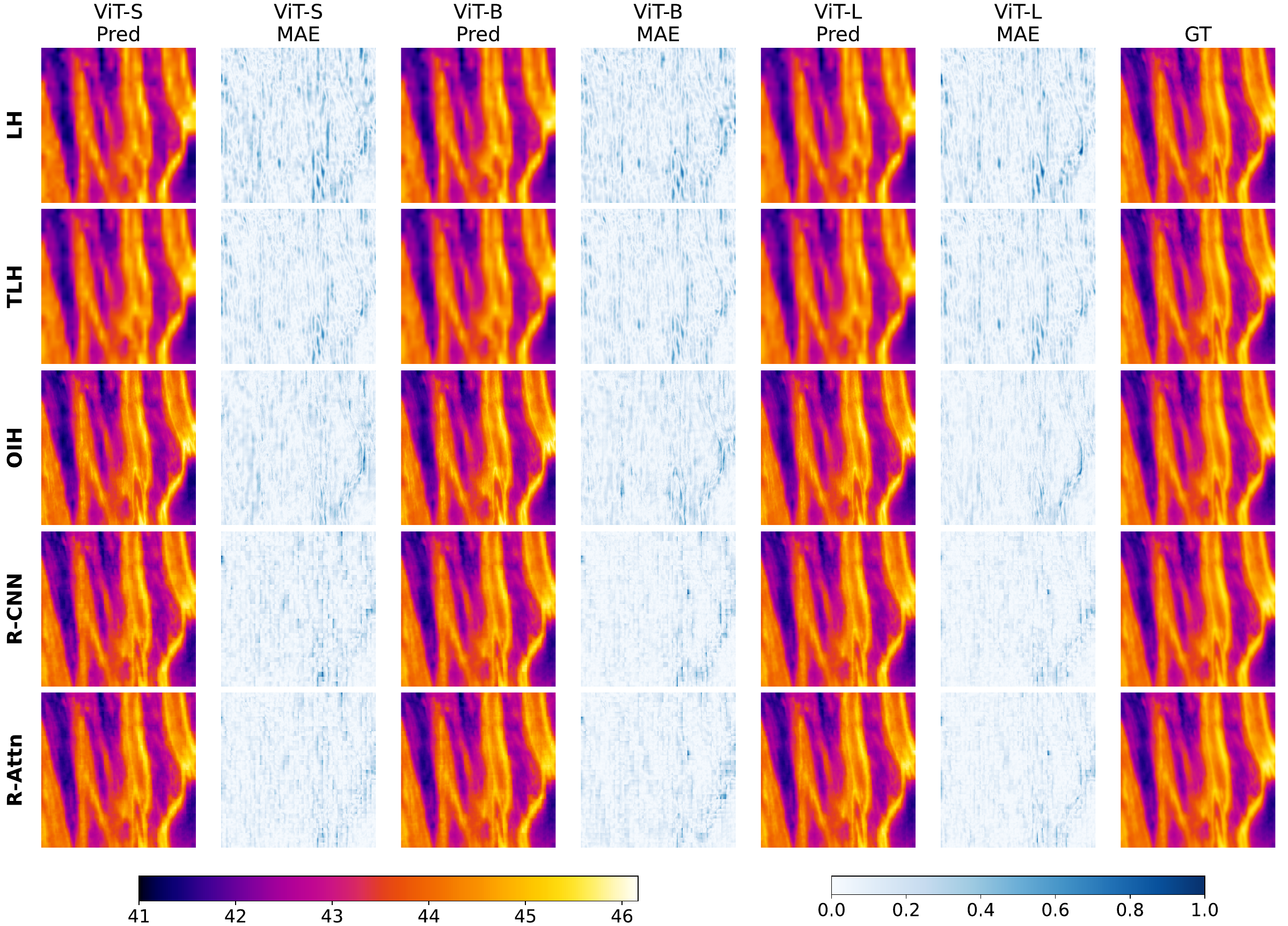}
     \caption{
     Qualitative results of different decoders based on DINOv3 backbones (ViT-S, ViT-B, ViT-L).
     }
    \label{fig:model_comparison}
\end{figure}

\paragraph{Results.}
Tab.~\ref{tab:master_performance_summary} presents the quantitative results of the proposed decoders across three distinct scales of the frozen DINOv3 backbone (ViT-S, ViT-B, and ViT-L), and Fig.~\ref{fig:model_comparison} shows the qualitative results.
Treating the LH as our primary baseline, the results demonstrate that both structural intelligence and backbone scaling are critical for high-fidelity thermodynamic surrogate modeling. 
The proposed Patch-to-Pixel Refiners (both R-CNN and R-Attn) outperform all baseline configurations across every evaluated metric and backbone scale.
Even when constrained by the smallest backbone variant (ViT-S), the R-CNN achieves an MAE of $0.0702$ and an RMSE of $0.0995$, surpassing the baseline LH configuration running on the ViT-L model.
The Refiner architectures demonstrate exceptional physical fidelity, pushing the strict accurate prediction index ($\delta_{01}$) to a near-perfect $99.84\%$ under the ViT-L configuration.

\subsubsection{Ablation Study}

To understand the mechanistic drivers behind these performance gaps, we break down these findings into three core physical and architectural insights.

\paragraph{Impact of Nonlinear Expressivity.}
As shown in Tab.~\ref{tab:master_performance_summary}, the linear projection of LH suffers from severe underfitting.
By simply introducing a single hidden layer and a GELU activation function, TLH achieves a dramatic performance leap, driving the ViT-S MAE down from $0.1150$ to $0.0840$. 
This indicates that while the frozen DINOv3 backbone provides rich structural features, a linear mapping lacks the expressive capacity to capture complex relationships between compressed semantic tokens and local energy density. 
A deeper token-wise expression is mandatory to correctly weight and transform the foundation model's features before mapping them to continuous thermal gradients.

\begin{figure}[t]
    \centering
    \includegraphics[width=1.0\textwidth]{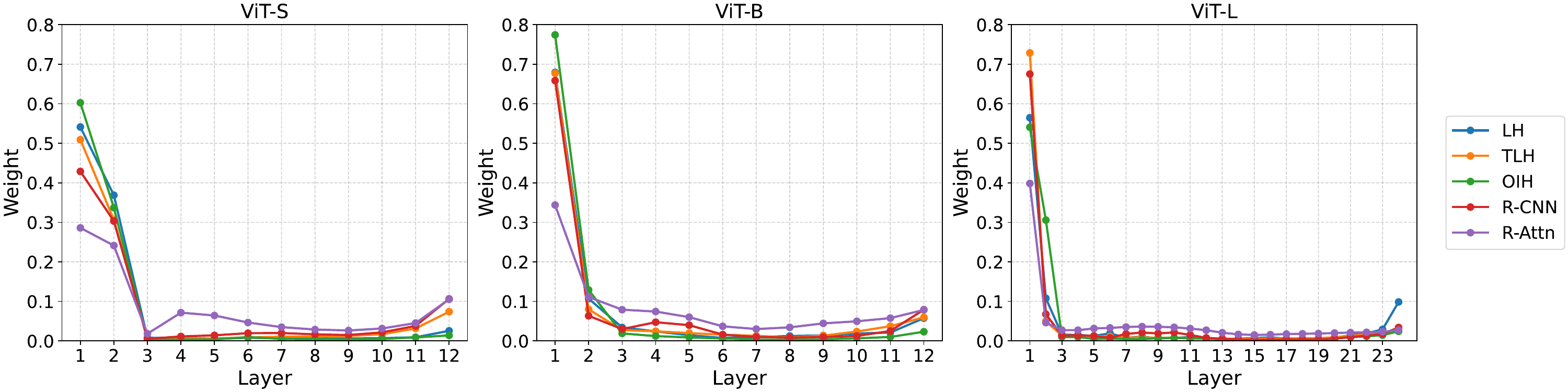}
     \caption{Distributions of learned softmax layer weights across different decoders and DINOv3 backbones (ViT-S, ViT-B, ViT-L).
     }
    \label{fig:layer_weights}
\end{figure}

\paragraph{Role of Geometry and High-Frequency Information.}
Pointwise upsampling architectures inherently lose high-frequency geometric boundaries due to the spatial dampening of bilinear interpolation. 
The OIH explicitly addresses this by introducing direct geometric shortcuts.
By concatenating the raw RGB image ($X$), OIH anchors its smooth thermodynamic predictions onto the sharp cellular boundaries of the physical wood sample, decreasing the ViT-S RMSE to $0.1266$ and establishing a stricter prediction profile. 
The R-CNN pushes this principle to its logical extreme by replacing interpolation with a dedicated patch-to-pixel upsampling network.
The critical necessity of this high-frequency geometry is empirically supported by the learned softmax layer weights (Fig.~\ref{fig:layer_weights}). 
Across all backbone sizes, the fusion mechanism consistently assigns the highest weights to the earliest transformer layers, proving that low-level structural and geometric features are universally required for establishing thermodynamic boundary conditions. 
To capture sharp thermal shifts across heterogeneous interfaces, a neural surrogate model must actively fuse smooth semantic tokens with these high-frequency spatial details to predict wood thermal responses.

\paragraph{Local vs. Global Contextual Refinement.} 
As seen in Tab.~\ref{tab:master_performance_summary}, when paired with the smallest backbone (ViT-S), R-Attn outperforms R-CNN (reducing the MAE from $0.0702$ to $0.0652$). 
This indicates that when the encoder's innate global receptive capacity is relatively constrained, adding an explicit global attention mechanism in the decoder helps resolve macroscopic thermodynamic dependencies before pixel unfolding. 
However, as the backbone scales to ViT-B and ViT-L, R-CNN overtakes R-Attn (e.g., ViT-L MAE is $0.0556$ for R-CNN compared to $0.0592$ for R-Attn).
Larger foundation models inherently construct a deeply integrated, globally coherent latent space.
In these regimes, imposing an additional layer of global attention introduces redundant complexity, whereas the strictly local R-CNN efficiently isolates the translation of existing global semantics into high-frequency spatial boundaries.
\paragraph{Effects of Backbone Encoder Scaling and Feature Utilization.}
Across all evaluated decoder topologies, scaling the frozen vision foundation model backbone from ViT-S to ViT-B and ViT-L yields consistent performance gains. 
This demonstrates that larger foundational models construct a more physically coherent and noise-resilient semantic latent space ($z_{\text{fuse}}$), containing finer multi-scale context regarding the wood's grain direction and density.
The scaling advantage of deep networks is fully realized by the patch-to-pixel refiners (R-CNN and R-Attn).

\section{Conclusion}
\label{sec:conclusion}

We present an end-to-end computational framework that effectively bridges the semantic and thermodynamic domains for high-resolution thermal analysis of wood.
By establishing optical intensity as a reliable geometric proxy for cellular solid volume fractions, we develop an automated FEM data engine to synthesize high-fidelity, pixel-level thermodynamic datasets.
To bypass the computational bottlenecks of these dense simulations, we subsequently design a fast neural surrogate framework leveraging DINOv3 foundation models to regress thermal responses directly from wood RGB images.
We demonstrate that a dedicated Patch-to-Pixel Refiner circumvents the spatial blurring inherent in standard upsampling, thereby anchoring thermal predictions to the complex wood anatomy. 
Our future work will combine 2D (image) and 3D (structure) micro-CT data to characterize the wood thermal behavior more accurately and explore using multi-modal large models to make our prediction framework more practical.

\bibliographystyle{ieeetr}
\bibliography{ref}

\newpage
\appendix
\section{Heat Transfer and Boundary Condition Verification}
\label{sec:verification}
\subsection{Thermodynamic Modeling along the Thickness Direction}
We assume that, for a single pixel, the thermal conductivity of $k_z$ is a function of $z$, i.e., $k_z = g(z)$ ($g(z)$ can be any function), and $f(z) = \frac{\partial}{\partial x} \left( k_{x} \frac{\partial T}{\partial x} \right) +
\frac{\partial}{\partial y} \left( k_{y} \frac{\partial T}{\partial y} \right)$.
We integrate $\frac{d}{d \xi} \left( g(\xi) \frac{d T}{d \xi} \right) = f(\xi)$ from $z$ to surface $L_z$
\begin{align}
    g(L_z) \left . \frac{d T} {d z} \right|_{z=L_{z}} - g(z) \frac{d T} {d z}  &= \int_{z}^{L_z} f(\xi) d\xi \label{eq:integrate_once}.
\end{align}
We denote the heat flux $-g(L_z) \left . \frac{d T} {d z} \right|_{z=L_{z}} = \left .-k_{\text{top}} \frac{d T} {d z} \right|_{z=L_{z}}$ as $Q$ and rearrange Eq.~\ref{eq:integrate_once}
\begin{align}
    \frac{dT}{dz} &= -\frac{Q}{g(z)} - \frac{1}{g(z)} \int_{z}^{L_z} f(\xi) d\xi, \\
    T_{\text{bed}} - T_{\text{top}} &= Q \int_{0}^{L_z} \frac{dz}{g(z)} + \int_{0}^{L_z} \frac{\int_{z}^{L_z} f(\xi) d\xi}{g(z)} dz.
\end{align}
We define an effective thermal conductivity $k_{\text{eff}}$ and the aggregated heat $\mathrm{\Phi}$ along the $z$ axis as
\begin{align}
    k_{\text{eff}} = \frac{L_z}{\int_{0}^{L_z} \frac{dz}{g(z)}}, \mathrm{\Phi} &= \int_{0}^{L_z} \frac{\int_{z}^{L_z} f(\xi) d\xi}{g(z)} dz, \\
    \frac{k_{\text{eff}}}{L_z} (T_{\text{bed}} - T_{\text{top}} - \mathrm{\Phi}) &= h(T_{\text{top}} - T_\infty). \label{eq:general_sol}
\end{align}
Based on Eq.~\ref{eq:general_sol}, the heat aggregation $\mathrm{\Phi}$ comes from the integral $f(z)$, which is shaped by the thermal conductivity profiles.
For a homogeneous case, assuming that $T_{\text{bed}}$ is uniform across the $xy$ plane, $\mathrm{\Phi} = 0$.
The thermal conductivity profile $g(z)$ is a constant function, i.e,  $k_z = k_{\text{eff}} = k_{\text{top}}$.
We reduce Eq.~\ref{eq:general_sol} to 
\begin{align}
    \label{eq:ana_linear0}
    k_{\text{top}} \frac{T_{\text{bed}} - T_{\text{top}}}{L_z} = h T_{\text{top}} - h T_{\infty}.
\end{align}

\subsection{Verification of the Heat Transfer Model and Boundary Conditions}
For four simultaneously measured samples (Oak Samples $\# 1 - 4$), spatial thermal responses are recorded under four distinct aluminum plate temperatures (30, 40, 50, and 60$^\circ \mathrm{C}$).
Each sample is cropped to a size of $200 \times 200$ after measurements.
To compress these 2D thermal responses into single representative values, three statistical estimators are applied across the $200 \times 200$ pixels, including the Overall Average (OA), the Median (ME), and the Interquartile Mean (IQM). 
Consequently, for each sample and each statistical estimator, exactly four data points are generated to estimate the heat transfer parameters ($h$ and $T_{\infty}$) given the fixed thermal conductivity $k_{\text{top}}$ (Eq.~\ref{eq:ana_linear0}).
Tab.~\ref{tab:linear_homo_params} shows the estimated parameters of ($h$ and $T_{\infty}$) using OA, ME, and IQM.
The results demonstrate this validity in two major ways: by matching expected physical ranges and by proving spatial consistency across measurements.
\begin{itemize}
    \item [1)]\textbf{Validation against Physical Ranges.}
    Across all four Oak samples and statistical estimators, the ambient temperature $T_{\infty}$ consistently operates within a range of $21.8$ to $25.0^\circ \mathrm{C}$. 
    This precisely reflects standard indoor ambient conditions, validating that the models correctly anchor to the physical environment.
    
    \item [2)] \textbf{Validation through Spatial Consistency.}
    The four Oak samples ($\# 1$, $\# 2$, $\# 3$, and $\# 4$) for either Side A or Side B are measured simultaneously under approximately identical environmental conditions.
    Because they are exposed to the same ambient environment at the same time, a physically accurate model should yield parameters that are closely grouped across all four locations.
    This strong spatial consistency across the four samples confirms that the internal medium is almost homogeneous, validating both the model and the boundary condition equations.
\end{itemize}

\begin{table}[t]
\centering
\caption{Estimated coefficient $h$ and ambient temperature $T_{\infty}$ across the four Oak samples.}
\label{tab:linear_homo_params}
\begin{tabular}{cccccc}
\toprule
\multirow{2}{*}{Estimator} & \multirow{2}{*}{Oak sample \#} & \multicolumn{2}{c}{Side A} & \multicolumn{2}{c}{Side B} \\
\cmidrule(lr){3-4} \cmidrule(lr){5-6}
& & $h[\text{W}\text{m}^{-2} \text{K}^{-1}]$ & $T_{\infty}$ [$^\circ \mathrm{C}$] & $h[\text{W}\text{m}^{-2} \text{K}^{-1}]$ & $T_{\infty}$ [$^\circ \mathrm{C}$] \\
\midrule
\multirow{4}{*}{OA}   & 1 & 22.271 & 22.171 & 16.987 & 24.344 \\
                              & 2 & 22.660 & 22.151 & 19.558 & 24.945 \\
                              & 3 & 22.414 & 21.945 & 17.057 & 24.286 \\
                              & 4 & 25.525 & 22.958 & 18.766 & 23.509 \\
\midrule
\multirow{4}{*}{ME}   & 1 & 22.244 & 22.220 & 16.670 & 24.253 \\
                              & 2 & 22.412 & 22.105 & 19.481 & 24.993 \\
                              & 3 & 21.963 & 21.754 & 17.385 & 24.274 \\
                              & 4 & 24.447 & 22.714 & 18.148 & 23.353 \\
\midrule
\multirow{4}{*}{IQM}  & 1 & 22.523 & 22.226 & 16.879 & 24.378 \\
                              & 2 & 23.001 & 22.219 & 19.485 & 24.867 \\
                              & 3 & 22.649 & 22.079 & 17.065 & 24.335 \\
                              & 4 & 25.584 & 23.010 & 18.590 & 23.357 \\
\bottomrule
\end{tabular}
\end{table}

\paragraph{Stability of Statistical Estimators.}
To ensure the reliability of the estimated parameters, the data are calculated using OA, ME, and IQM.
The variance among these three estimators is small, often within fractions of a degree or coefficient unit.
Because IQM, which removes the upper and lower $25 \%$ of the spatial data, aligns nearly perfectly with OA, it can be concluded that the steady-state experimental data across the $200 \times 200$ pixels are normally distributed and highly stable. 
The estimated parameters are deeply tied to the physical core of the system rather than being skewed by transient experimental noise, edge effects, or sensor artifacts.

\section{Derivation of Visual-Thermal Modeling}
\label{sec:modeling}
We assume that $k_x$, $k_y$, and $k_{z}$ are locally uniform within a small micro-neighborhood. 
For a single pixel $(x,y)$, $k_z$ remains constant along the $z$ direction but varies across the $xy$-plane. 
We can therefore factor the thermal conductivities out of the spatial derivatives
\begin{equation}
k_{x} \frac{\partial^2 T}{\partial x^2} + k_{y} \frac{\partial^2 T}{\partial y^2} + k_{z} \frac{\partial^2 T}{\partial z^2} = 0.
\label{eq:3d_anisotropic1}
\end{equation}
Defining the anisotropy ratios as constants, $\alpha = k_x/k_z$ and $\beta = k_y/k_z$, and setting $\alpha = \beta$ for simplicity, Eq.~\ref{eq:3d_anisotropic1} reduces to
\begin{equation}
\alpha \nabla^2 T + \frac{\partial^2 T}{\partial z^2} = 0 \quad \Leftrightarrow \quad \frac{\partial^2 T}{\partial z^2} = -\alpha \nabla^2 T,
\label{eq:3d_simpler1}
\end{equation}
where $\nabla^2 = \frac{\partial^2}{\partial x^2} + \frac{\partial^2}{\partial y^2}$ denotes the 2D Laplace operator. 
Because the wood sample operates strictly within a thin-plate regime ($L_z =$ 3 mm), we approximate the testbed temperature distribution ($z=0$) by performing a second-order Taylor series expansion of the temperature field $T(x,y,z)$ along the $z$-axis, anchored at the top surface ($z=L_z$)
\begin{equation}
T(x,y,0) = T(x,y,L_z) - L_z \left. \frac{\partial T}{\partial z} \right|_{z=L_z} + \frac{L_z^2}{2} \left. \frac{\partial^2 T}{\partial z^2} \right|_{z=L_z} - \mathcal{O}(L_z^3).
\label{eq:taylor}
\end{equation}
The first and second derivatives are  evaluated at the bottom and top boundaries
\begin{equation}
\left . \frac{\partial T} {\partial z} \right|_{z=L_z} = - \frac{h}{k_{\text{top}}} \left( T_{\text{top}} - T_{\infty} \right), \quad
\left. \frac{\partial^2 T}{\partial z^2} \right|_{z=L_z} = -\alpha \nabla^2 T_{\text{top}}.
\end{equation}
Defining the temperature difference as $\Delta T(x,y) = T_{\text{bed}}(x,y) - T_{\text{top}}(x,y)$, the true spatial Laplacian of the surface temperature expands to $\nabla^2 T_{\text{top}} = \nabla^2 T_{\text{bed}} - \nabla^2 \Delta T$. Substituting these boundary conditions back into the Taylor expansion and isolating $\Delta T$ yields
\begin{equation}
\Delta T - \frac{\alpha L_z^2}{2} \nabla^2 \Delta T = \frac{L_z h}{k_{\text{top}}} \left( T_{\text{top}} - T_{\infty} \right) - \frac{\alpha L_z^2}{2} \nabla^2 T_{\text{bed}}.
\label{eq:corrected_poisson}
\end{equation}
Because it is modulated by the squared thin-plate parameter ($L_z^2 \approx$ $10^{-6} \text{ m}^2$), the Laplacian term $\frac{\alpha L_z^2}{2} \nabla^2 T_{\text{bed}}$ is infinitesimally small and approaches $0$, simplifying Eq.~\ref{eq:corrected_poisson} to
\begin{equation}
\Delta T - \frac{\alpha L_z^2}{2} \nabla^2 \Delta T = \frac{L_z h}{k_{\text{top}}} \left( T_{\text{top}} - T_{\infty} \right).
\label{eq:corrected_poisson_new}
\end{equation}

To solve Eq.~\ref{eq:corrected_poisson_new} analytically, we define the diffusion parameter as $\lambda^2 = \frac{\alpha L_z^2}{2}$ and the surface convective heat flux as $q = h(T_{\text{top}} - T_{\infty})$.
Applying the inverse resolvent operator, $(1 - \lambda^2 \nabla^2)^{-1}$, to both sides isolates the temperature difference $\Delta T$.
For small values of $\lambda$ inherent to the thin-plate assumption, the resolvent is asymptotically equivalent to the continuous spatial evolution operator $e^{\lambda^2 \nabla^2}$. 
Treating $\lambda^2$ as a pseudo-time variable, the Green's function for this diffusion process is mathematically defined as
\begin{equation}
\Phi(x,y;\lambda) = \frac{1}{4 \pi \lambda^2} \exp\left(-\frac{x^2+y^2}{4\lambda^2}\right).
\label{eq:greens_function}
\end{equation}
Applying the evolution operator to a spatial field is exactly equivalent to convolving the field with this Green's function. 
By comparing Eq.~\ref{eq:greens_function} to a standard 2D Gaussian kernel $G_{\sigma}$, we find they are perfectly identical when the variance is mapped as $\sigma^2 = 2\lambda^2 = \alpha L_z^2$. 
Thus, the exact analytical solution to the thermal response becomes
\begin{equation}
\Delta T(x,y) \approx G_{\sigma} \ast \left( \frac{L_z q}{k_{\text{top}}(x,y)} \right).
\label{eq:blurred_resistance1}
\end{equation}

Given $k_{\text{top}}(x,y) = k_{\max} \left[ 1 - I(x,y) \right]$, inverting Eq.~\ref{eq:blurred_resistance1} and applying a first-order Taylor approximation ($\left[ G_{\sigma} \ast (k_{\text{top}})^{-1} \right]^{-1} \approx G_{\sigma} \ast k_{\text{top}}$) yields the inverse temperature difference
\begin{equation}
\frac{1}{\Delta T(x,y)} \approx \frac{1}{L_z q} \left[ G_{\sigma} \ast \left( k_{\max} \left[ 1 - I(x,y) \right] \right) \right].
\label{eq:inverse_T}
\end{equation}
By defining a single system parameter $\eta = k_{\max} / (L_z q)$, we rearrange Eq.~\ref{eq:inverse_T} to
\begin{equation}
\frac{1}{\Delta T} \approx \eta \left[ G_{\sigma} \ast (1-I) \right].
\label{eq:w_parameter}
\end{equation}
We approximate the macroscopic heat flux $q$ using 1D Fourier's law over the bulk wood sample, $q \approx \bar{q} = \bar{k}_{\text{bulk}} \frac{\Delta \bar{T}}{L_z}$, where $\bar{k}_{\text{bulk}}$ is the macroscopic bulk thermal conductivity and $\Delta \bar{T}$ is the spatially averaged temperature difference. 
Comparing the parameters yields $k_{\max} = \eta \bar{k}_{\text{bulk}} \Delta \bar{T}$. 
For simplicity, we non-dimensionalize the macroscopic thermal potential by normalizing $\bar{k}_{\text{bulk}} \Delta \bar{T} = 1$. 
The system parameter $\eta$ thus directly represents the effective intrinsic thermal conductivity ($\eta = k_{\max}$). 
Consequently, we can synthesize high-resolution maps of localized thermal conductivity directly from novel RGB images using
\begin{equation}
k_{\text{top}}(x,y) = \eta \left[ 1 - I(x,y) \right].
\label{eq:final_synthesis}
\end{equation}

\begin{figure}[t]
    \centering
    \includegraphics[width=1.0\textwidth]{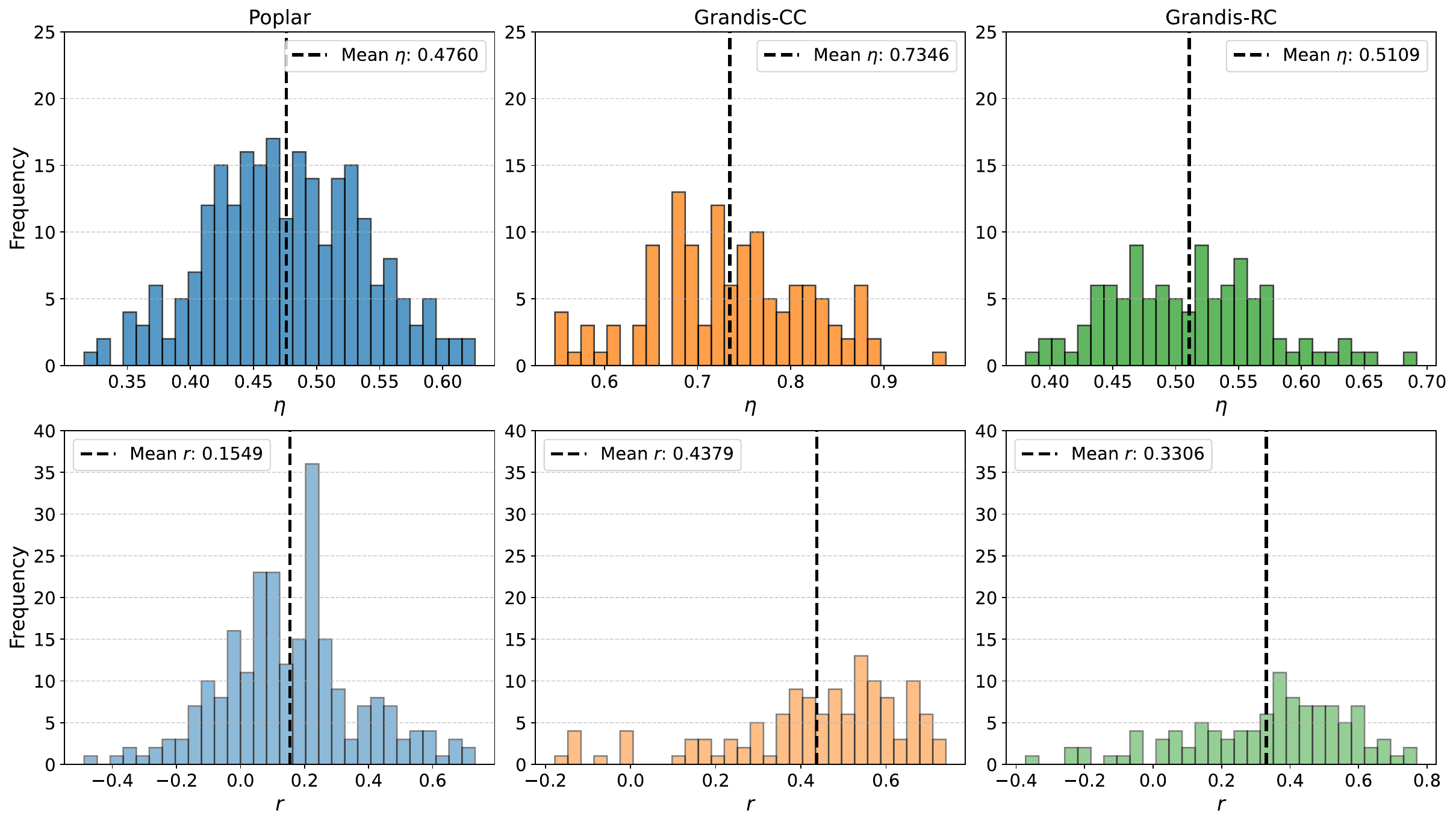}
    \caption{Distributions of the system parameter, $\eta$, and the Pearson correlation coefficient, $r$, are evaluated on each sample across Poplar, Grandis-CC, and Grandis-RC datasets. 
    }
    \label{fig:histograms}
\end{figure}

\paragraph{Distributions of the System Parameter $\eta$.}
To evaluate sample-to-sample stability, $\eta$ and the Pearson correlation coefficient ($r$) are regressed independently for every sample (Fig.~\ref{fig:histograms}). 
The extracted $\eta$ values form well-defined distributions around distinct dataset means.
The correlation coefficients further validate the physical assumption, demonstrating strong spatial mappings across datasets.
\label{sec:hist_w}

\section{Derivation of the Weak Formulation for the 3D Heat Equation}
\label{sec:derivation_weak_form}

The steady state 3D anisotropic heat equation without internal heat generation is
\begin{equation}
\nabla \cdot (\mathbf{K} \nabla T) = 0, \quad \text{in } \Omega,
\end{equation}
where $\Omega$ represents the 3D computational domain ($L_x \times L_y \times L_z$) and $\mathbf{K}$ is the anisotropic thermal conductivity tensor. 
The domain boundary $\partial \Omega = \Gamma_{\text{bottom}} \cup \Gamma_{\text{top}} \cup \Gamma_{\text{sides}}$ is subject to Dirichlet ($T = T_{\text{bed}}$ at $z=0$), Robin ($-(\mathbf{K} \nabla T) \cdot \mathbf{n} = h(T - T_{\infty})$ at $z=L_z$), and Neumann ($-(\mathbf{K} \nabla T) \cdot \mathbf{n} = 0$ on the lateral sides) boundary conditions, where $\mathbf{n}$ denotes the outward-pointing unit normal vector.
To formulate the weak problem, we define the Sobolev space $V = H^1(\Omega)$. 
We construct the trial function space $V_T$ and test function space $V_0$, where functions in $V_0$ must vanish on the Dirichlet boundary to preserve the known physical constraints
\begin{equation}
V_T = \{ T \in V : T = T_{\text{bed}} \text{ on } \Gamma_{\text{bottom}} \}, \quad V_0 = \{ v \in V : v = 0 \text{ on } \Gamma_{\text{bottom}} \}.
\end{equation}

Multiplying the strong form by an arbitrary test function $v \in V_0$ and integrating over the entire spatial domain $\Omega$ yields
\begin{equation}
\int_{\Omega} v [\nabla \cdot (\mathbf{K} \nabla T)] \, d\Omega = 0.
\end{equation}
Applying the product rule for divergence and the Divergence Theorem transforms the volumetric integral into a boundary integral over the closed surface $\partial \Omega$
\begin{equation}
\int_{\Omega} (\mathbf{K} \nabla T) \cdot \nabla v \, d\Omega = \int_{\partial \Omega} v (\mathbf{K} \nabla T) \cdot \mathbf{n} \, ds.
\label{eq:variational_base}
\end{equation}
Because $v = 0$ on $\Gamma_{\text{bottom}}$ (Dirichlet) and the physical heat flux is strictly zero on $\Gamma_{\text{sides}}$ (Neumann), these two integrals vanish.
We substitute the Robin boundary condition into the remaining integral on the top surface
\begin{align}
\int_{\Omega} (\mathbf{K} \nabla T) \cdot \nabla v \, d\Omega &= \int_{\Gamma_{\text{top}}} v [-h(T - T_{\infty})] \, ds,\\
\int_{\Omega} (\mathbf{K} \nabla T) \cdot \nabla v \, d\Omega + \int_{\Gamma_{\text{top}}} h T v \, ds &= \int_{\Gamma_{\text{top}}} h T_{\infty} v \, ds.
\end{align}

\section{Thermal Conductivity Distribution along the Thickness Direction}
\label{sec:thermal_conductivity_distributions}

We consider three typical distributions of thermal conductivity along the thickness direction for wood samples, including \textit{Uniform}, \textit{Linear}, and \textit{Random}. 
These distributions are defined at the pixel level in the $xy$ plane over the two-dimensional spatial domain and describe how the thermal conductivity $k_{z,i,j}(z)$ varies along the thickness direction. 
\paragraph{Uniform.}
The uniform thermal conductivity refers to the situation where, for each pixel in the $xy$-plane, the thermal conductivity remains constant throughout the entire thickness of the wood sample
\begin{equation}
k_{z, i, j}(z) = k_{z, i, j}, \forall z \in [0, L_{z}].
\end{equation}

\paragraph{Linear.}
The linear thermal conductivity represents a continuous and gradually varying thermal property of the wood sample along the thickness direction
\begin{equation}
\label{eq:lin_graded}
k_{z,i,j}(z) = \left( 1 - \frac{z}{L_{z}} \right) k_{\text{bot},i,j} + \frac{z}{L_{z}} k_{\text{top},i,j},
\end{equation}
where $k_{\text{bot}}$ and $k_{\text{top}}$ represent the thermal conductivities at the bottom and top surfaces, respectively.

\paragraph{Random.}
The random thermal conductivity describes a stratified wood structure composed of $M$ discrete layers along the thickness direction, each with a uniform thermal conductivity
\begin{equation}
k_{z, i, j}(z) = k_{m, i, j}, z \in (z_{m-1}, z_{m}), m=\{1, \cdots, M\},
\end{equation}
where $\{z_{0}=0, z_{1}, \dots, z_{M}=L_{z}\}$ define the layer interfaces, and $\{k_{1,i,j}=k_{\text{bot}}, \dots, k_{M,i,j}=k_{\text{top}}\}$ denote the thermal conductivities of individual layers.

\end{document}